%% file: neurips_2023.tex
\documentclass{article}

\usepackage[preprint]{neurips_2023}

\usepackage[utf8]{inputenc} 
\usepackage[T1]{fontenc}    
\usepackage{hyperref}       
\usepackage{url}            
\usepackage{booktabs}       
\usepackage{amsfonts}       
\usepackage{nicefrac}       
\usepackage{microtype}      
\usepackage{xcolor}         
\usepackage{graphicx}
\usepackage{todonotes}
\usepackage{subcaption}
\usepackage{multirow}
\usepackage{pgfplots}
\usepackage{makecell}
\usepackage{tikz}
\usepackage{colortbl}
\usepackage{wrapfig}
\usepackage{enumitem,kantlipsum}

\input{math_commands.tex}

\input{figures}

\title{FineQuant: Unlocking Efficiency with Fine-Grained Weight-Only Quantization for LLMs}

\author{%
  Young Jin Kim\thanks{Equal contribution.} \\
  Microsoft \\
  \texttt{youki@microsoft.com} \\
  \And
  Rawn Henry\footnotemark[1] \\
  NVIDIA \\
  \texttt{rhenry@nvidia.com} \\
  \AND
  Raffy Fahim \\
  Microsoft \\
  \texttt{raffybekheit@microsoft.com} \\
  \And
  Hany Hassan Awadalla \\
  Microsoft \\
  \texttt{hanyh@microsoft.com} \\
}

\begin{document}

\maketitle

\begin{abstract}
Large Language Models (LLMs) have achieved state-of-the-art performance across various language tasks but pose challenges for practical deployment due to their substantial memory requirements. Furthermore, the latest generative models suffer from high inference costs caused by the memory bandwidth bottleneck in the auto-regressive decoding process. To address these issues, we propose an efficient \texttt{weight-only} quantization method that reduces memory consumption and accelerates inference for LLMs.
To ensure minimal quality degradation, we introduce a simple and effective heuristic approach that utilizes only the model weights of a pre-trained model. This approach is applicable to both Mixture-of-Experts (MoE) and dense models without requiring additional fine-tuning.
To demonstrate the effectiveness of our proposed method, we first analyze the challenges and issues associated with LLM quantization. Subsequently, we present our heuristic approach, which adaptively finds the granularity of quantization, effectively addressing these problems. Furthermore, we implement highly efficient GPU GEMMs that perform on-the-fly matrix multiplication and dequantization, supporting the multiplication of \texttt{fp16} or \texttt{bf16} activations with \texttt{int8} or \texttt{int4} weights.
We evaluate our approach on large-scale open source models such as OPT-175B and internal MoE models, showcasing minimal accuracy loss while achieving up to 3.65 times higher throughput on the same number of GPUs.
\end{abstract}

\section{Introduction}

Large Language Models (LLMs) have proven their efficacy in various language tasks by increasing the number of trainable parameters and pre-training models on large-scale data to be used in different downstream tasks \citep{devlin2018bert, radford2018improving, liu2019roberta, raffel2020exploring}. With the advancement of distributed large-scale training methods \citep{Shazeer2018MeshTensorFlowDL, Rasley2020DeepSpeedSO, Ren2021ZeROOffloadDB, baines2021fairscale} and large-scale data collection \citep{raffel2020exploring, hoffmann2022training}, models have grown even larger and achieved state-of-the-art performance with increased capacity to learn, demonstrating the capability for in-context learning \citep{brown2020language, zhang2022opt, chowdhery2022palm} that can be used for various language tasks even without updating parameters for specific tasks. \cite{zhang2022opt}

However, deploying such large models comes with a significant cost which increases proportionally with the model size. Model size growth has increased several orders of magnitude over the last few years (1,588 times larger from \texttt{BERT large} - 340 million to \texttt{PaLM} 540 billion)\citep{devlin2018bert, chowdhery2022palm}, and without improving inference efficiency, inference cost and latency will rise dramatically.

Quantization is a compression technique that reduces model size and speeds up inference by approximating floating-point numbers with smaller precision numbers. Numerous studies have demonstrated the effectiveness of quantization in accelerating neural network model inference \citep{rodriguez2018lower, stock2019and, Choukroun2019LowbitQO, Gholami2022ASO}, particularly in natural language generation, such as machine translation \citep{kim2019research, Aji2020CompressingNM, Fan2021TrainingWQ, park2022nuqmm, kim2022says} and natural language understanding tasks \citep{kim2020fastformers}. However, it is still under-explored how weight-only quantization can be effectively utilized in the context of large language models. Also, the existing methods introduce complex and costly procedures such as additional Quantization Aware Training (QAT) and/or calibration on additional data. Otherwise, they compromise either speed or accuracy. 
To more effectively solve the challenge, we focus on simple weight-only quantization method that requires no additional training in this study because it has multiple advantages - (i) the accuracy could be maintained well because its underlying numerical computation is done in floating-point precision which is more accurate. As a result, we can effectively push the precision to very low bit-ranges. (ii) it could be used for various hardware and GPU architectures without needing specific hardware instructions dealing with low-bit multiplications. (iii) it can avoid expensive additional training steps. Then, the key research questions are how to effectively exploit this low-bit quantization without losing accuracy and how to efficiently implement a GEMM which accepts different types on modern GPUs.

In this paper, we make the following contributions:
\begin{enumerate}[wide, labelwidth=!, labelindent=0pt]
\item Extensive Analyses of Quantization Behaviors: We provide comprehensive analyses of the quantization behaviors on Language Model Models (LLMs). We investigate the impact of applying low-bit quantization (down to 3-bits) on LLM accuracy.

\item Fine-Grained Quantization Algorithm: We propose a fine-grained quantization algorithm that incorporates group-wise quantization and adaptive selection of granularity. This approach helps preserve the original floating-point precision accuracy even when there is loss due to quantization.

\item Highly Efficient GPU Kernels: We implement highly efficient GPU kernels and conduct a thorough performance analysis, considering different batch sizes and context lengths. This analysis allows us to identify the optimal utilization of the proposed approach on real GPUs.

\item Accelerated Inference with Large-Scale Models: We demonstrate the effectiveness of the proposed method by applying it to a large-scale open-source dense transformer model called OPT. With its 175 billion parameters and internal MoE models utilizing optimized GPU kernels, our method enables deployment of the 175 billion parameter model on only 2 GPUs, resulting in a significant reduction of overhead and cost by 64\%. Moreover, our method achieves 3.65 times higher throughput on the same number of GPUs.
\end{enumerate}

These contributions collectively advance the understanding of quantization behaviors in LLMs, propose an effective quantization algorithm, optimize GPU implementation, and demonstrate the practical benefits in terms of reduced resource requirements and improved inference throughput.

\section{Background - Challenges of Quantizing LLMs}

\subsection{Fundamental challenges of inferencing generative LLMs}

\textbf{Increased communication overhead. }
We must issue an all reduce after each attention and FFN block when doing inference with tensor parallelism. While technologies such as NVLink and NCCL greatly accelerate GPU to GPU communication, it is desirable to use as few GPUs as possible to minimize this overhead.


\textbf{Large weights with small activations. }
The increase in the model size causes the matrix multiplies in the decoding phase of LLMs to be bottlenecked by memory bandwidth. The weights typically dominate the memory traffic as the activations tend to only have a few tokens once the context has been used to generate the KV attention caches. As the number of parameters increase, the amount of data that must be moved from HBM to the GPU cores increases which places even more pressure on the memory subsystem. In modern processors, compute is much faster than memory so it is desirable to reduce the memory bottleneck.

Given those observations, it is critical to reduce the memory footprint.

\subsection{Quantization challenges}
Quantization is an active research topic to accelerate inference and reduce the memory footprint of LLMs. However, there are still many challenges remaining, and especially there is no single method which can maintain the accuracy and improve the efficiency at the same time without introducing complex procedures to convert and execute an inference.

\textbf{It is hard to maintain good accuracy when applying quantizaiton on LLMs. }
It is known that naive quantization methods could significantly degrade the accuracy compared to the original models' \citep{frantar2022gptq}. One reason for this is outliers in the activation based on the previous studies \citep{Dettmers2022LLMint88M, xiao2022smoothquant}. \citet{Dettmers2022LLMint88M, xiao2022smoothquant} proposed methods to mitigate this issue by handling the outliers separately in floating-point arithmetic or by shifting the multiplier to the model weights from the activations.

\textbf{It is difficult to achieve high efficiency.}
Even if some algorithms could maintain the accuracy of the original floating-point models, it is also non-trivial to get efficient implementation of the proposed method in reality. This requires special kernel implementations on GPUs. For example, \citet{Dettmers2022LLMint88M} could achieve a good accuracy with quantization, but the efficiency improvement was marginal. Also, OPTQ \citet{frantar2022gptq} does not provide efficient inference kernels other than batch size 1. However, we note that our efficient GPU kernels can be used with weights quantized by OPTQ., allowing one to benefit from the speed of our kernels and the accuracy of OPTQ.

\textbf{Added complexity to solve the problem.}
To overcome the issues of accuracy drop and inefficiency of runtimes, there have been several studies proposed. Those approaches require expensive and complex procedures to achieve the goal, especially with target task specific dataset for the calibration. \citet{Yao2022ZeroQuantEA} uses additional knowledge distillation steps to recover the accuracy drop from the quantization. \citet{park2022nuqmm} uses binary coding quantization and it performs iterative numerical optimization to find the best binary coding scheme for a given model and a task which is non-trivial. \citet{frantar2022gptq} uses Optimal Brain Quantization (OBQ) to maintain the accuracy of the original floating-point model which shuffles the model weights based on the approximated second-order Hessian matrix information. All of those approaches have introduced non-trivial and dataset specific algorithmic procedures. Especially, the cost of those algorithms grows together with the size of the base models.

In this work, our goal is to find a scalable, accurate and efficient quantization method without introducing additional cost of model conversion.

\section{Designing Quantization Methods for LLMs - Adaptive Fine-grained Quantization}
\label{sec:quant_methods}
This section delves into the phenomenon observed in LLM quantization, specifically focusing on potential issues that can lead to quality degradation, particularly in relation to the quantization range. We thoroughly examine these issues and explore potential strategies to mitigate them while ensuring effective control over the quantization range. Building on our analysis, we propose a heuristic algorithm designed to automatically determine the appropriate quantization range.

\subsection{Quantization methodology: basic settings}

\textbf{Uniformity of quantization}

We conducted experiments involving two quantization techniques that focus on the uniformity of the quantized range. Firstly, we employed linear quantization, which uniformly maps quantized integer values to their corresponding original float values. Secondly, we explored log-based quantization, inspired by \citet{Aji2020CompressingNM}, where both integer and float ranges are mapped in a logarithmic scale. In both cases, we applied column-wise quantization to assess the impact of quantization uniformity on model accuracy. Detailed formulations for those two techniques are described in Appendix A.

Figure \ref{fig:expert FFNs log_opt_s vs linear} illustrates the performance comparison between two quantization techniques applied to FFN layers using low bits. For 3 and 4 bits, both techniques exhibit similar performance. However, with 2-bit quantization, log-scale quantization shows a significant decrease in accuracy. Considering these observations and the computational simplicity, we opt to use uniform quantization for all subsequent experiments.

\begin{wrapfigure}{r}{0.45\textwidth}
    \centering
    \vskip -0.15in
    \includegraphics[width=0.45\textwidth]{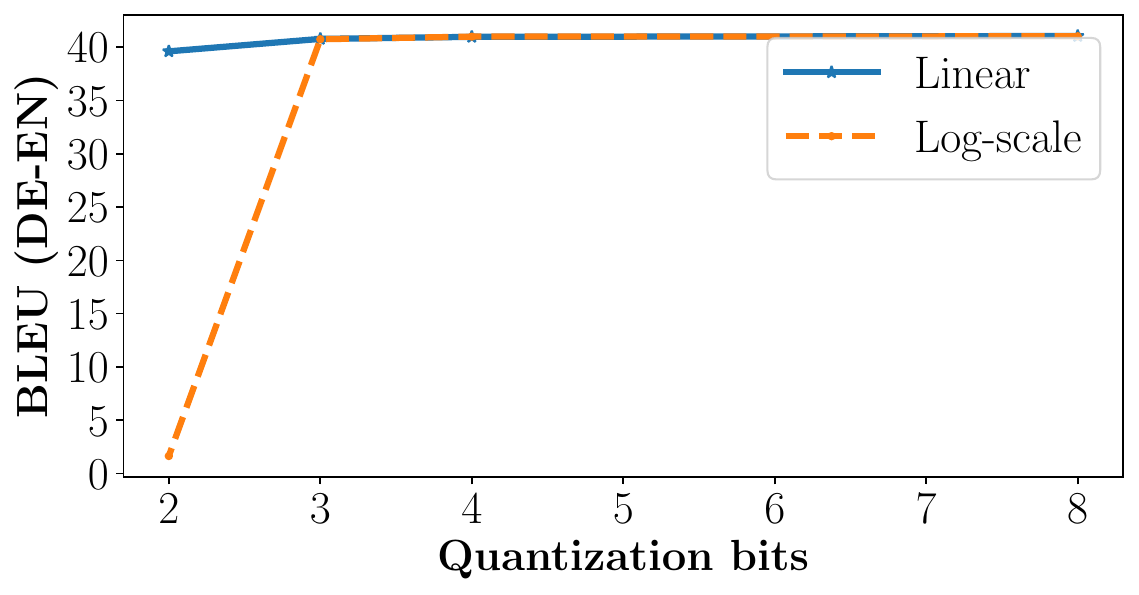}
    \vskip -0.1in
    \caption{\centering A comparison of how the quality of the model, as measured by BLEU, changes when quantizing with different precisions using different quantization methods.}
    \label{fig:expert FFNs log_opt_s vs linear}
\end{wrapfigure}

\textbf{Symmetricity - numerical distribution of model weights}

\begin{figure}[!h]
    \centering
    \begin{subfigure}[b]{0.495\textwidth}
        \centering
        \includegraphics[width=0.8\textwidth]{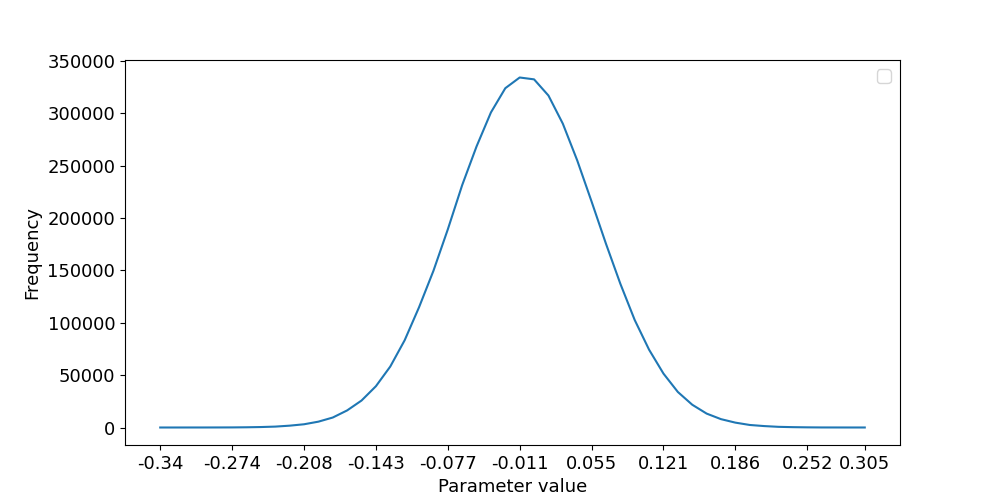}
        \vskip -0.1in
        \caption{\centering Example expert weight distribution\hspace{\textwidth}(layer 6, FFN 2, expert 15)}
        \label{fig:encoder-expert-15-layer-6-fc2-weights}
        \end{subfigure}
    \begin{subfigure}[b]{0.495\textwidth}
        \centering
        \includegraphics[width=0.8\textwidth]{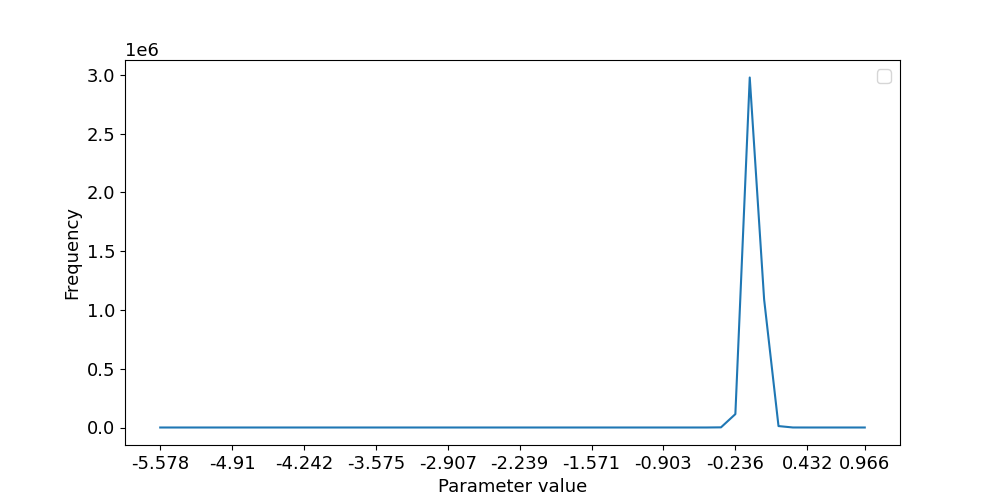}
        \vskip -0.1in
        \caption{\centering Example FFN weight distribution\hspace{\textwidth}(layer 7, FFN 2)}
        \label{fig:encoder-fc2-layer-7-weights}
    \end{subfigure}
    \caption{A comparison of example weight distributions from MoE and dense FFN layers.}
    \label{fig:weight-distribution}
\end{figure}

\label{sec:weight-distribution}
In order to determine the most appropriate quantization approach, we have conducted further analysis on the weight parameter distribution across various layers. Figure~\ref{fig:weight-distribution} presents example distributions of model weights, which generally exhibit a normal distribution centered around zero. However, in some cases, outliers can distort the weight distribution, potentially leading to an inaccurate quantization range. Based on our observations and considering implementation efficiency, we choose to employ symmetric quantization around zero.

\subsection{Granularity of quantization}
\label{sec:groupwise}
Considering the design choices made earlier in this section, the granularity of quantization emerges as the most crucial component of the quantization algorithm. For the sake of efficient computation and reduced memory consumption, it is typical to have 1 quantization scale per tensor or 1 quantization scale for each column in the tensor. However, to maintain a close approximation of the original numerical values with the quantized values, it is desirable to have smaller groups of parameters sharing scales. This is necessary because outliers in the distribution have the potential to significantly skew the data, leading to decreased quantization precision, especially for smaller numerical values.

\subsubsection{Catastrophic collapse of model performance}
\begin{figure}[!h]
    \centering
    \begin{subfigure}[b]{0.495\textwidth}
        \centering
        \includegraphics[width=0.9\textwidth]{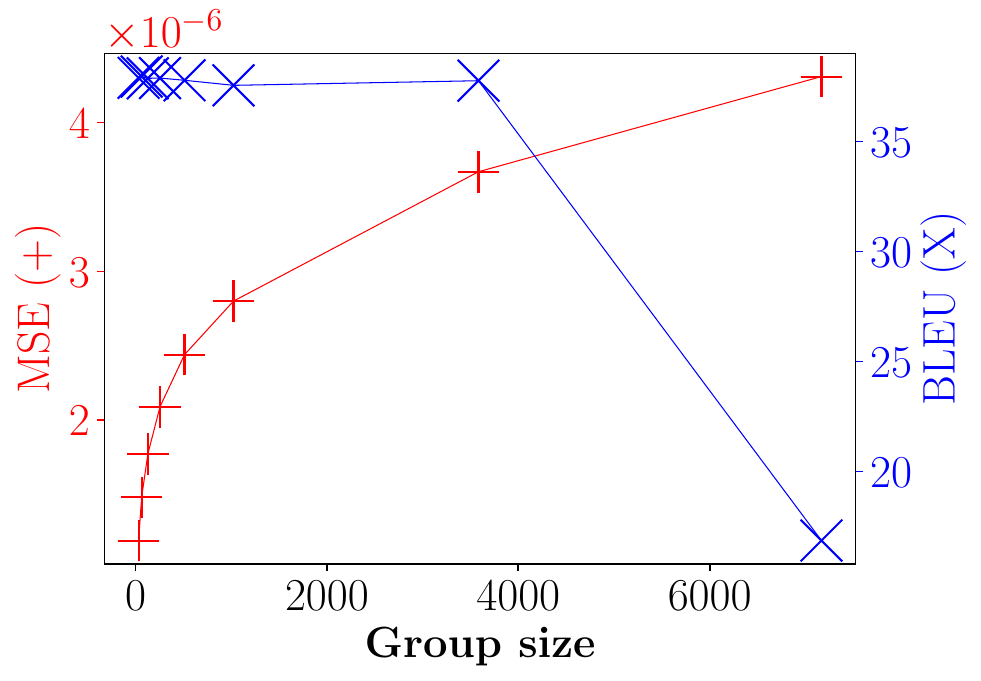}
        \caption{\centering MSE and BLEU changes with quantization group sizes.}
         \label{fig:mse-bleu}
    \end{subfigure}
    \begin{subfigure}[b]{0.495\textwidth}
        \centering
        \includegraphics[width=0.85\textwidth]{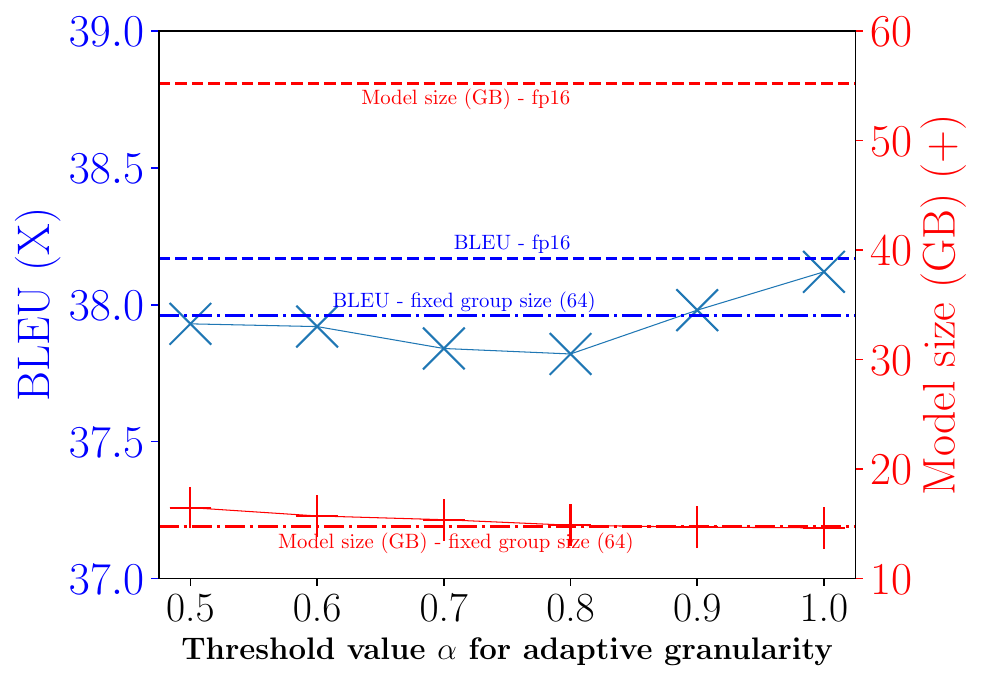}
        \caption{\centering BLEU score and model size comparison with adaptive group quantization with reference lines of fp16 and fixed group size (64). X-axis represents threshold value $\alpha$ of adaptive fine-grained quantization.}
        \label{fig:group-threshold}
    \end{subfigure}
    \caption{Impact analyses of quantization granularity on translation accuracy of OPT-30B.}
    \label{fig:group-analysis}
\end{figure}

Throughout our observations, we have noted a significant decline in performance when employing matrix-wise quantization compared to column-wise quantization across various layers, as demonstrated in Appendix B. Consequently, column-wise quantization serves as the baseline for our experiments. However, even with column-wise quantization, we have encountered instances of catastrophic collapse in LLM performance, particularly when certain outliers exist in the model weights. Figure \ref{fig:mse-bleu} depicts the relationship between the Mean Squared Error (MSE) of quantized values and the translation BLEU scores as we modify the group size in the OPT 30B model. While increasing granularity leads to a gradual rise in MSE values, the model quickly loses its capability in terms of task BLEU score beyond a certain point. Consequently, it is crucial to determine the optimal granularity for each matrix to preserve the task performance while maximizing the size of the parameter groups which share scales.


\subsection{Adaptive fine-grained quantization}
Upon further investigation into the catastrophic failure of a quantized model, we have discovered that the failure could be rectified by adjusting the granularity of four specific matrices out of the 288 quantized matrices. Merely increasing the granularity of these four matrices by a factor of two allowed for the recovery of over 94\% of the lost accuracy. Based on this observation, we have developed a simple heuristic-based method to assign varying granularity to different model weight matrices.

In the process of quantizing a matrix, we start from the column-wise quantization and compute the range of the values that must be quantized. We then halve the quantization group size and compute the range of each group. If for any group, $\frac{new\_range}{old\_range} > \alpha$ we halve the quantization group size again. We repeat this process until the quantization range differences between two granularities becomes smaller than $\alpha$. Figure \ref{fig:group-threshold} illustrates the impact of adaptive group size on BLEU scores and model sizes in gigabytes (GB). With the adaptive fine-grained quantization approach, there is only a marginal 0.1\% difference in BLEU score, while the model size is reduced to a mere 26\% of the original FP16 model size.

\section{Experiments}
\subsection{Experimental setup}
Our latency and throughput experiments are conducted using NVIDIA A100 SXM4 GPUs inside a Docker container running Ubuntu 20.04 and CUDA 11.8. All code is compiled using nvcc 11.8.89 and gcc/g++ 9.3. To carry out the experiments, we use a modified version of FasterTransformer \footnote{\url{https://github.com/NVIDIA/FasterTransformer}} v5.3. The weight-only quantization kernels for per-column quantization are already open source.

\textbf{Task and datasets.}
For the dense models, we utilize various open-source language tasks, including LAMBADA, HellaSwag, PiQA, WinoGrande, OpenBookQA, RTE, COPA from the lm-evaluation harness \citep{eval-harness}, as well as WMT machine translation task (WMT16 German and English)\footnote{\url{https://statmt.org/wmt16/}}.

For the MoE models, we use a multilingual machine translation task that covers 10 language translation directions from and into English covering German (de), French (fr), Italian (it), Spanish (es), Dutch (nl), and English (en). We use a 128,000 sub-word vocabulary, built with the sentencepiece library\footnote{\url{https://github.com/google/sentencepiece}}. The number of training sentences is included in Appendix E. To measure the accuracy of the models, we utilized sacrebleu ~\footnote{\url{https://github.com/mjpost/sacrebleu}} on the detokenized output.

\textbf{Dense model architecture.}
For the dense model experiments, we utilize various open-source large language models that share a similar architecture, which consists of decoder-only with multiple transformer layers. To evaluate the accuracy of these models, we include GPT-2-XL (1.5B) \citep{radford2019language}, OPT (13B and 30B) \citep{zhang2022opt}, and OPT-IML (Max 30B and Max 175B) \citep{iyer2022opt}. The number of model parameters ranges from 1.5 billion to 175 billion. The detailed number of layers and hidden dimensions can be found in the original papers.

\textbf{MoE model architecture.}
For our MoE model experiments, we utilize internal pre-trained MoE models (5.3B) with a few modifications to the transformer model architecture \citep{Vaswani2017AttentionIA}. These modifications encompass the following: (i) a deep encoder consisting of 24 transformer layers and a shallow decoder comprising 12 transformer layers, (ii) adoption of Transformer with Untied Positional Encoding (TUPE) proposed in \citet{Ke2021RethinkingPE} instead of the conventional sinusoidal positional embedding, and (iii) implementation of pre-layer normalization from \citet{Xiong2020OnLN}.
For the MoE models, we employ top-1 learned gating from \citet{fedus2021switch} and an MoE layer with 32 experts at every other layer, specifically the even-numbered layers, as utilized in \citet{lepikhin2020gshard, fedus2021switch, kim2021scalable}. Additionally, we apply jittering noise, balancing loss (ratio of 0.01) \citep{lepikhin2020gshard, fedus2021switch} to more uniformly distribute expert utilization and gating dropout (0.2) \citep{liu2022gating} to prevent overfitting and improve regularization.

\textbf{GPU kernel implementations.}
We utilized the kernel implementations developed by \citet{kim2022says}, which rely on CUTLASS to create efficient kernels for fused dequantization and matrix multiplication. These kernels can process either FP16 or BF16 activations, a vector of scales of the same data type as the activation, and int8 or int4 weights. The kernels dequantize the weights to match the data type of the activation and perform floating-point tensor core math. The final output of the kernel is also of the same data type as the input activation. These kernels are available as open source code in FasterTransformer. To support multiple scaling factors for each column, we extended these kernels to process a matrix of scales, enabling us to implement int4 block quantization kernels. We set the block size to 64 for all performance analyses below, since it matches the K tile size of our fused gemm + dequantize kernels.

In compute-bound cases such as an encoder or the context creation phase of GPT, the conversions from integer to float bottlenecks our kernels, rather than tensor core math. As a result, our weight-only quantization GEMMs slower than equivalent FP16xFP16 GEMMs in compute bound cases but offer significant speedup in memory bound cases as seen in Figure \ref{fig:optspeedups}. We argue that this kernel is useful because:
\begin{enumerate}[wide, labelwidth=!, labelindent=1pt]
    \item Large language models (LLMs) usually spend a lot more time in the memory-bound decoding phase than in the compute-bound context creation phase, especially when the output sequence length is long. 
    \item LLMs are typically served with small batch sizes in most practical cases, which puts significant pressure on the memory system during matrix multiplication as the weights need to be read from the GPU's HBM. However, our kernel utilizes int4 compression, which reduces the number of bytes needed to load the weights by up to 4X. The overhead of loading the scales is small, even for block quantization with block size 64 as shown in Figure~\ref{fig:optspeedups}.
\end{enumerate}

\textbf{Quantization Method.}
All quantization experiments have one scaling factor for each column of the weight matrix, unless a block size $B$ is specified. In that case, each contiguous block of $B$ elements in a given column has its own scaling factor. This means we have multiple scaling factors per column.

\subsection{Dense model performance results}

\subsubsection{Accuracy}
Table~\ref{tab:dense-acc-result} presents the impact of quantization on various natural language tasks using different models. The results show that, in general, 8-bit weight-only quantization does not significantly affect the accuracy compared to fp16. This is observed across different language tasks, indicating that the models produce similar outputs. However, 4-bit quantization with column-wise granularity leads to some degradation in accuracy due to outliers in the weight distribution, as discussed in Section~\ref{sec:groupwise}. To recover the accuracy, we adopt a group-wise quantization strategy, which shows similar accuracy to the original fp16.

We also show similar experiments for OPT-IML for machine translation. Table~\ref{tab:optiml-mt} shows the accuracy numbers with different bit quantization on OPT-IML 30B and 175B models. With a group-wise quantization approach, the models could preserve the accuracy while quantizing down to 4-bit and 3-bit for some parts.

\subsubsection{Microbenchmarks}
To understand how our weight-only quantization accelerates the matrix multiplies, we collect micro-benchmarks from OPT-13B and OPT-30b and present the results in Figure~\ref{fig:optspeedups}. We find that the matrix multiplies can be accelerated by up to 2.5X for those models when the number of tokens in the activation is small. This is typically the case for the auto-regressive part of LLMs which tends to dominate the overall run-time. 

\subsubsection{End to End Benchmarks}
\begin{figure}
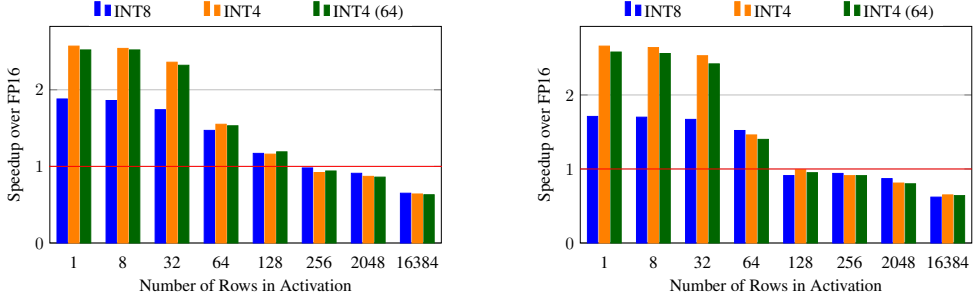

\begin{minipage}[c]{0.5\linewidth}
\resizebox{6cm}{!}{%
\chartOptOneThree
}
\end{minipage}
\hfill
\begin{minipage}[c]{0.5\linewidth}
\resizebox{6cm}{!}{%
\chartOptThreeZero
}
\end{minipage}%

\caption{Demonstrates the speed up over FP16 on only the matrix multiplies for OPT-13B (left) and OPT-30B (right) on a single GPU. We measure the performance of the QKV Projection, Attention Output, FFN1 and FFN2 matrix multiplies and compare our CUTLASS FP16 x INT GEMM against cuBLAS performing FP16 x FP16 GEMM. We report the geometric mean of the speedups across those 4 GEMMs while varying the number of rows in the activation (which represents batch\_size $\times$ sequence\_length). We highlight that when the number of rows is small (such as the decoding phase of GPT), we achieve up to 2.5X GEMM speedup when doing int4 quantization with block size 64.} \label{fig:optspeedups}

\end{figure}

We construct Table~\ref{table:opt175bspeed} as a reference to compute end to end times for different input and output lengths for OPT-175B on 8, 4 and 2 GPUs. Our table shows that the context phase slows done which is primarily due to running on fewer GPUs. Additionally, our weight-only quantization kernels have some slowdown for compute bound cases. However, we show that the time per decoder step is typically within 20 \% of FP16 despite using 2X or 4x fewer GPUs. The per-token latency does not scale with the number of GPUs since fewer GPUs need to communicate and our kernels provide significant acceleration (as shown in Table~\ref{fig:optspeedups}) in the decoder phase.

Table~\ref{table:opt175bthroughput} shows end to end times (constructed from Table~\ref{table:opt175bspeed}) and associated throughput increases. To calculate the throughput increase, we assume the original FP16 model was sharded across 8-GPUs within a single node and that same node is used to serve INT8 or INT4 models. We measure the throughput per node by assuming that the model is replicated twice on the node for INT8 and 4 times for INT4 (64) and that requests are served to the independent model instances concurrently. We highlight that our compression technique allows serving 4 instances of OPT-175B on a single A100 node with 8 GPUs.

\begin{table*}[t]
\caption{Accuracy of various models with low-bit weight only quantization on different natural language tasks. We also include the perplexity on the wikitext dataset for each model. We note that with int4 per-column for OPT-30B actually performs worse than FP16 for OPT-13B. Using block quantization (with block size 64) improves the accuracy by 2.3 \% over just using per-column}
\setlength{\tabcolsep}{4.5pt}
\label{tab:dense-acc-result}
\begin{center}
\begin{scriptsize}
\begin{tabular}{c|c|c|c|c|c|c|c|c|c|c}
\hline
{Model type} & {Precision} & {LAMBADA} & {HellaSwag}   & {PiQA}  & {WinoGrande}  & {OBQA}  & {RTE}  & {COPA}  & {\bf Average $\uparrow$} & {\bf Wikitext $\downarrow$} \\
\hline
\multirow{4}{*}{GPT2-XL} & fp16  & 51.1\% & 40.0\% & 70.7\% & 58.2\% & 22.4\% & 52.3\% & 73.0\% & 52.5\% & 20.4 \\\cline{2-11}
& int8  & 51.1\% & 40.0\% & 70.7\% & 58.3\% & 22.6\% & 52.7\% & 73.0\% & 52.6\% & 20.4 \\\cline{2-11}
& int4 (64)  & 49.3\% & 39.6\% & 70.7\% & 58.4\% & 20.6\% & 50.9\% & 74.0\% & 51.9\% & 20.9 \\\cline{2-11}
& int4  & 47.5\% & 37.4\% & 69.4\% & 57.1\% & 19.4\% & 51.9\% & 73.0 \%& 50.8\% & 21.7 \\
\hline
\multirow{4}{*}{OPT-13B} & fp16  & 68.6\% & 52.5\% & 75.9\% & 65.0\% & 26.6\% & 58.1\% & 86.0\% & 61.8\% & 11.5 \\\cline{2-11}
& int8  & 68.5\% & 52.4\% & 76.0\% & 65.4\% & 27.\%2 & 57.0\% & 86.0\% & 61.8\% & 11.5 \\\cline{2-11}
& int4 (64)  & 67.4\% & 50.7\% & 75.6\% & 65.4\% & 25.8\% & 59.2\% & 84.0\% & 61.2\% & 12.0 \\\cline{2-11}
& int4  & 65.5\% & 50.2\% & 75.5\% & 64.8\% & 26.4\% & 56.0\% & 85.0\% & 60.5\% & 12.8 \\
\hline
\multirow{4}{*}{OPT-30B} & fp16  & 71.5\% & 54.3\% & 77.6\% & 68.2\% & 30.2\% & 57.4\% & 82.0\% & 63.0\% & 10.7 \\\cline{2-11}
& int8  & 71.4\% & 54.3\% & 77.6\% & 67.9\% & 30.2\% & 58.1\% & 82.0\% & 63.0\% & 10.7 \\\cline{2-11}
& int4 (64)  & 69.9\% & 53.4\% & 77.5\% & 67.3\% & 30.0\% & 56.0\% & 83.0\% & 62.4\% & 11.1 \\\cline{2-11}
& int4  & 69.5\% & 51.9\% & 75.8\% & 66.3\% & 26.8\% & 54.9\% & 79.0\% & 60.1\% & 11.6 \\
\hline
\hline
\end{tabular}
\vskip -0.1in
\end{scriptsize}
\end{center}
\end{table*}

\begin{table}[t]
\caption{Perplexity using LM Eval Harness and FasterTransformer. OPT 66B suffers from catastrophic collapse with INT4 per column quantization, but recovers with block quantization with a size of 64.}
\label{tab:wikitextoptlarge}
\centering
\scriptsize
\begin{tabular}{c|cccc|cccc}\hline
    \multirow{2}{*}{\textbf{Dataset}} & \multicolumn{4}{c|}{\textbf{OPT 66B}} & \multicolumn{4}{c}{\textbf{OPT 175B}} \\
    & FP16 & INT8 per col & INT4 per col & INT4 (64) & FP16 & INT8 per col & INT4 per col & INT4 (64) \\\hline        
    Wikitext & 10.15 & 10.15 & 143.16 & 10.66 & 9.08 & 9.08 & 11.08 & 9.84 \\
    \bottomrule
\end{tabular}
\end{table}

\begin{table}[t]
\vskip -0.25in
    \caption{We show the time taken to construct the context and the time per decoder step for OPT-175B on 8, 4 and 2 GPUs using our different weight-only quantization schemes. The numbers for int4 per-column quantization are similar to int4 (64) so they are omitted. The compute bound context creation phase is up to 3.5X slower when using INT4 block quantization, but running on 4x fewer GPUs. For INT8, it is up to 1.9X slower but runs on 2X fewer GPUs. In addition, the time per decoder step is typically within 20\% of FP16 despite using 2X for 4X fewer GPUs with weight-only quantization. End to end times for different numbers of generated tokens can be estimated from this table by identifying the batch size and input length of interest and computing: context\_time + num\_generated\_tokens * time\_per\_decoder\_step. The batch sizes and sequence lengths shown are the maximum sizes that could fit in GPU memory.}\label{table:opt175bspeed}

\begin{center}
\begin{scriptsize}
    \begin{tabular}{c|c|cc|cc|cc}\hline
        \multirow{2}{*}{\textbf{\thead{Batch \\Size}}} & \multirow{2}{*}{\textbf{\thead{Input\\ length}}} & \multicolumn{2}{c|}{\textbf{\thead{FP16 (8 GPUs)}}} & \multicolumn{2}{c|}{\textbf{\thead{INT8 (4 GPUs)}}} & \multicolumn{2}{c}{\textbf{\thead{INT4 (64) (2 GPUs)}}} \\
          &     & \thead{Context \\ time (ms)} & \thead{Avg time \\ per decoder \\ step (ms)} & \thead{Context \\ time (ms)} & \thead{Avg time \\ per decoder \\ step (ms)} & \thead{Context \\ time (ms)} & \thead{Avg time \\ per decoder \\ step (ms)}\\\hline        
        1  & 128 &  60  & 40 & 76   & 38 & 121  & 43\\\hline
        2  & 128  & 82  & 41 & 134  & 38 & 226  & 42\\\hline
        4  & 128  & 148 & 41 & 283  & 38 & 431  & 43\\\hline
        8  & 128  & 272 & 41 & 468  & 40 & 835  & 45\\\hline
        12 & 128  & 372 & 42 & 743  & 41 & 1173 & 48\\\hline
        16 & 128  & 491 & 42 & 890  & 42 & 1627 & 49\\\hline
        32 & 128  & 935 & 44 & 1776 & 47 & 3261 & 58 \\\hline
        1  & 512  & 148 & 42 & 280  & 40 & 427  & 44\\\hline
        2  & 512  & 273 & 43 & 470  & 40 & 838  & 45\\\hline
        4  & 512  & 493 & 43 & 892  & 41 & 1637 & 46\\\hline
        8  & 512  & 939 & 43 & 1784 & 43 & 3291 & 49\\\hline
        1  & 1024 & 271 & 41 & 465  & 39 & 829  & 44\\\hline
        2  & 1024 & 498 & 42 & 899  & 39 & 1648 & 46\\\hline
        4  & 1024 & 945 & 42 & 1795 & 42 & 3307 & 50\\
        \bottomrule
    \end{tabular}
\vskip -0.2in
    \end{scriptsize}
\end{center}
\end{table}

\begin{table}[t]
\setlength{\tabcolsep}{3.0pt}
\vskip -0.2in
\caption{Shows the throughput improvement for batch 1 on a 8-GPU node for different input and output lengths. We assume that the model is replicated twice on the node for INT8 weight-only quantization and 4 times for INT4 (64) weight-only quantization. We show the throughput increase relative to FP16 in parentheses next to through-puts for INT8 and INT4. The table is constructed using data from Table~\ref{table:opt175bspeed}.}\label{table:opt175bthroughput}
\begin{center}
\begin{scriptsize}
\vskip -0.1in
\begin{tabular}{ c|c|c|c|c }
    
\textbf{\thead{Input \\Length}} & \textbf{\thead{Output \\Length}} & \textbf{\thead{FP16 throughput \\ per 8 GPU node \\ (generated tokens per sec)}} & \textbf{\thead{INT8 throughput \\ per 8 GPU node \\ (generated tokens per sec)}} & \textbf{\thead{INT4 (64) throughput \\ per 8 GPU node \\ (generated tokens per sec)}} \\\hline
128  & 32  & 24 & 49  (2.04$\times$) & 85  (3.54$\times$) \\ \hline
128  & 128 & 25 & 52  (2.08$\times$) & 91  (3.64$\times$) \\ \hline
512  & 32  & 21 & 41  (1.95$\times$) & 69  (3.29$\times$) \\ \hline
512  & 128 & 23 & 47  (2.04$\times$) & 84  (3.65$\times$) \\ \hline
1024 & 32  & 20 & 37  (1.85$\times$) & 57  (2.85$\times$) \\ \hline
1024 & 128 & 23 & 47  (2.04$\times$) & 79  (3.43$\times$) \\
 
\end{tabular}
\end{scriptsize}
\end{center}
\end{table}

\begin{table*}[t]
\vskip -0.15in
\caption{Accuracy of OPT-IML models (30B and 175B) with various weight only quantization settings on machine translation tasks. The BLEU score is used as a metric and higher number represent a better result. The group size for the group-wise quantization is specified together with quantization bits.}
\label{tab:optiml-mt}
\begin{center}
\begin{scriptsize}
\vskip -0.1in
\begin{tabular}{c|c|c|c|c}
\hline
{\bf Model type} & {\bf Attention (group)} & {\bf Others (group)} & {\bf WMT 2016 German to English} &   {\bf Model Footprint (GB)}   \\
\hline
& fp16  & fp16 & \textbf{38.20} & 55.21 \\\cline{2-5}
& int8 (7,168)  & int8 (7,168) & \textbf{38.20} & 27.66 \\\cline{2-5}

& int4 (16)  & int4 (16) & 38.10 & 17.32  \\\cline{2-5}
& int4 (64)  & int4 (64) & 37.96 & 14.73 \\\cline{2-5}
\multirow{3}{*}{OPT-IML Max 30B} & int4 (7,168)  & int4 (7,168) & 16.86 & 13.88 \\\cline{2-5}
& int4 (adaptive)  & int4 (adaptive) & 38.12 & 14.62 \\\cline{2-5}

& int3 (64)  & int4 (64) & 37.75 & 13.87 \\\cline{2-5}

& int4 (64)  & int3 (64) & 37.06 & 12.15 \\\cline{2-5}

& int3 (16)  & int3 (16) & 37.57 & 13.87 \\\cline{2-5}
& int3 (64)  & int3 (64) & 36.95 & 11.29 \\\cline{2-5}
& int3 (7,168)  & int3 (7,168) & 0.00 (degenerate) & 10.43 \\\cline{2-5}

\hline
 & fp16  & fp16 & 41.14 & 324.16 \\\cline{2-5}
& int8 (12,288)  & int8 (12,288) & \textbf{41.18} & 162.18 \\\cline{2-5}
\multirow{2}{*}{OPT-IML Max 175B} & int4 (64)  & int4 (64) & 40.86 & 86.23 \\\cline{2-5}
& int4 (12,288)  & int4 (12,288) & 0.00 (degenerate) & 81.19 \\\cline{2-5}
& int3 (64)  & int4 (64) & 40.93 & 81.16 \\\cline{2-5}
& int4 (64)  & int3 (64) & 37.02 & 71.04 \\\cline{2-5}
\hline
\hline
\end{tabular}
\end{scriptsize}
\end{center}
\end{table*}

\subsection{MoE model performance results}
We evaluate the performance of our weight-only quantization method on an MoE model and report the results in Table \ref{tab:mose-result}. We investigate the impact of different quantization precisions, ranging from 8-bit to 3-bit. Due to the robustness of the MoE FFN layers, the model's accuracy is preserved quite well even with 3-bit and 4-bit precision, when compared to the original fp16 accuracy.

\begin{table*}[!h]
\vskip -0.15in
\caption{Accuracy of MoE models with quantization. Speed-up comparison is presented in Figure~\ref{fig:moespeed}. The optimized kernels are implemented for 8-bit and 4-bit precisions.}
\label{tab:mose-result}
\begin{center}
\begin{scriptsize}
\vskip -0.1in
\begin{tabular}{c|cccccc}
\hline
\bf Model type & \bf Precision & {\bf BLEU} ($\Delta$ BLEU compared to fp16)  & {\bf Size} (X times compared to fp16) \\
\hline
\multirow{4}{*}{MoE 5.3B} & fp16  & {46.35 (0.0)}  & 1.00X  \\\cline{2-4}
& int8  & {46.34 (-0.01)}  & 0.55X \\\cline{2-4}
& int4  & 46.18 (-0.17)  & 0.32X  \\\cline{2-4}
& int3  & 46.01 (-0.34)  & 0.26X  \\\cline{2-4}
\hline
\hline
\end{tabular}
\end{scriptsize}
\end{center}
\end{table*}

\begin{figure}[h]
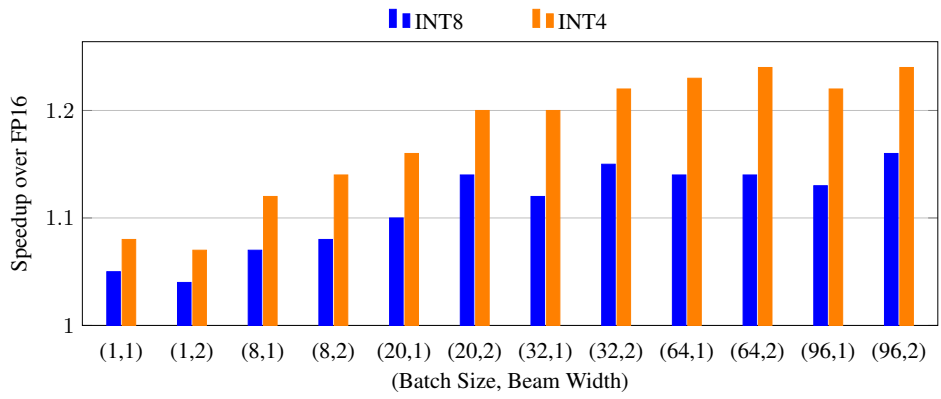

    \centering
    \vskip -0.05in
    \resizebox{0.9\linewidth}{!}{%
    \chartMoeSpeedup
    }
    \vskip -0.1in
    \caption{MoE model speed-up with quantization methods.}
    \label{fig:moespeed}
\end{figure}

Figure~\ref{fig:moespeed} shows the end-to-end speed improvements with various batch size with 8-bit and 4-bit quantization.

\section{Conclusions and Limitations}
This paper presents a method for accelerating large language models through the use of low-bit quantization. The proposed weight-only quantization technique demonstrates promising results in compressing very large models with up to 175 billion parameters, while still maintaining accuracy. To address the issue of outliers affecting the quantized weight distribution, fine-grained quantization is employed.

Despite its strengths, the study does have a few limitations. Firstly, optimized GPU kernels are only implemented for group size 64. However, we plan to expand support for any power of 2 group size greater than 16. Secondly, the performance benchmarking is conducted solely on A100 GPUs, so the speed improvements may vary on different GPU architectures. Lastly, the proposed method does not leverage integer instructions even when they are available. These limitations suggest potential directions for future research.

One particularly promising avenue for future work involves exploring the accuracy and efficiency of using int8 activations and int4 weights with integer scales for fine-grained quantization. This approach has the potential to further enhance the efficiency of the models.

\clearpage

\bibliography{anthology,custom}
\bibliographystyle{acl_natbib}

\clearpage

\appendix
\section{Quantization method formulation}
\label{app:uniformity}
\textbf{Linear quantization with absolute maximum.}
We used linear quantization with absolute maximum as the main method. Given a matrix $\mA$ and $b$ bits, this method encodes $\mA$ as follows:

\begin{center} \label{eq1:linquant}
$\displaystyle \vs_{j} = \frac{2 \times \max(|\mA_{:,j}|)}{2^{b}-1}$ \\
$\displaystyle \mQ_{:,j}= \integer(\frac{\mA_{:,j}}{\vs_{j}})$ \\

\end{center}

Here, $s$ is the scaling factor, which can be chosen per channel, as shown, or per the whole tensor. At inference time, the quantized $\mQ$ is dequantized back to $\mA^{'}$ with the scaling factor $\vs$ as follows:

\begin{center}
$\displaystyle \mA^{'}{:,j}=\mQ{:,j} \times \vs_{j}$ \\
\end{center}

\textbf{Log-scale quantization.}
Another quantization method we experimented is log-scale quantization where $1$ bit is kept for the sign and (${b-1}$) bits are used to encode the log-scaled values. Given a matrix $\mA$, the quantization formula is as follows:
\begin{center}
$\displaystyle \mP = sign({\mA})$ \\
$\displaystyle \mT =clip(\frac{|\mA|}{s}, 1, 2^{1-2^{b-1})}$ \\
$\displaystyle \mQ = \lceil log_2(\frac{2}{3}\mT) \rceil$  \\
\end{center}
where $s$ can be chosen in two ways, either (i) the absolute maximum or (ii) the optimal value to minimize the mean squared error (MSE) between the quantized and original values which is described in \citet{Aji2020CompressingNM}. We use the second algorithm which we observe a better accuracy with the quantization. At inference time, the quantized weight values are dequantized based on the formula as follows:
\begin{center}
$\displaystyle \mA^{'}= \mP\times s \times 2^\mQ$ \\
\end{center}

Figure \ref{fig:expert FFNs log_opt_s vs linear} shows the performance comparison of two quantization methods.

\section{Channel-wise vs matrix-wise quantization}
\label{app:chan-mat}
\begin{figure}[h]
    \centering
    \includegraphics[width=0.5\textwidth]{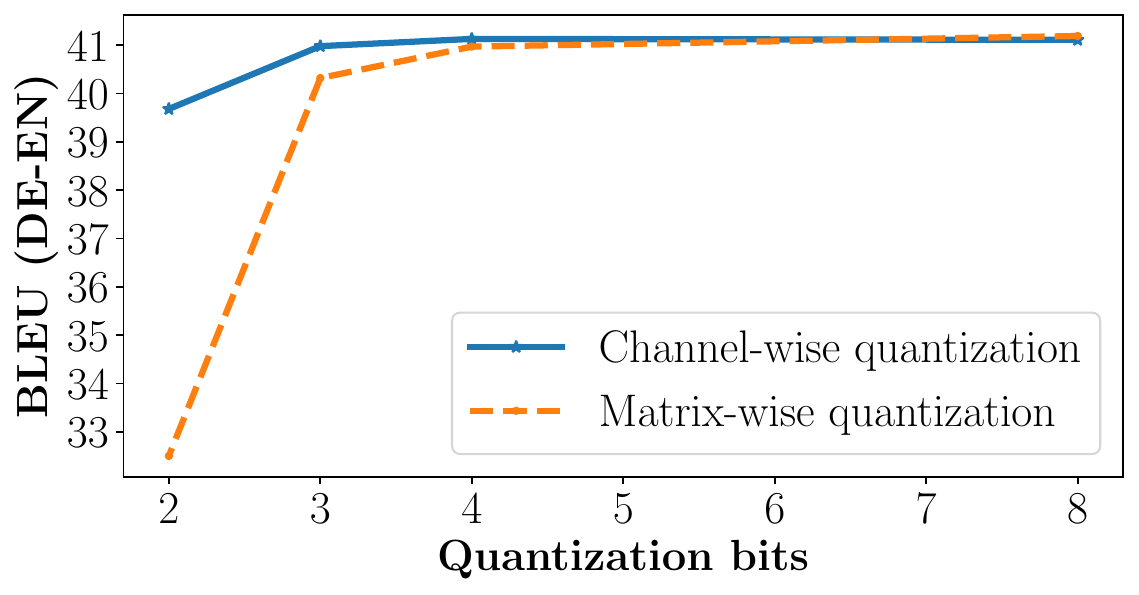}
    \vskip -0.1in
    \caption{Linear quantization of expert FFNs with channel-wise and matrix-wise scaling factors.}
    \label{fig:linear_channel_vs_tensor_expert_ffns}
\end{figure}

Scaling factors are calculated by the quantization algorithm and stored in half precision floating-point (fp16) numbers to dequantize the matrices with. These factors can be chosen on the channel scale or the whole matrix scale. As shown in figure \ref{fig:linear_channel_vs_tensor_expert_ffns}, channel-wise quantization gives quite higher scores than tensor-wise especially for low precision. Additional parameters to store channel-wise scaling factors is small, because only one value is needed for a channel and less than 1\% of total parameters in a matrix. Therefore, we use channel-wise quantization for all the quantization experiments.

\section{Quantization of different layers in a dense model}
\label{app:dense-layers}
\begin{figure}[h]
    \centering
    \includegraphics[width=0.5\textwidth]{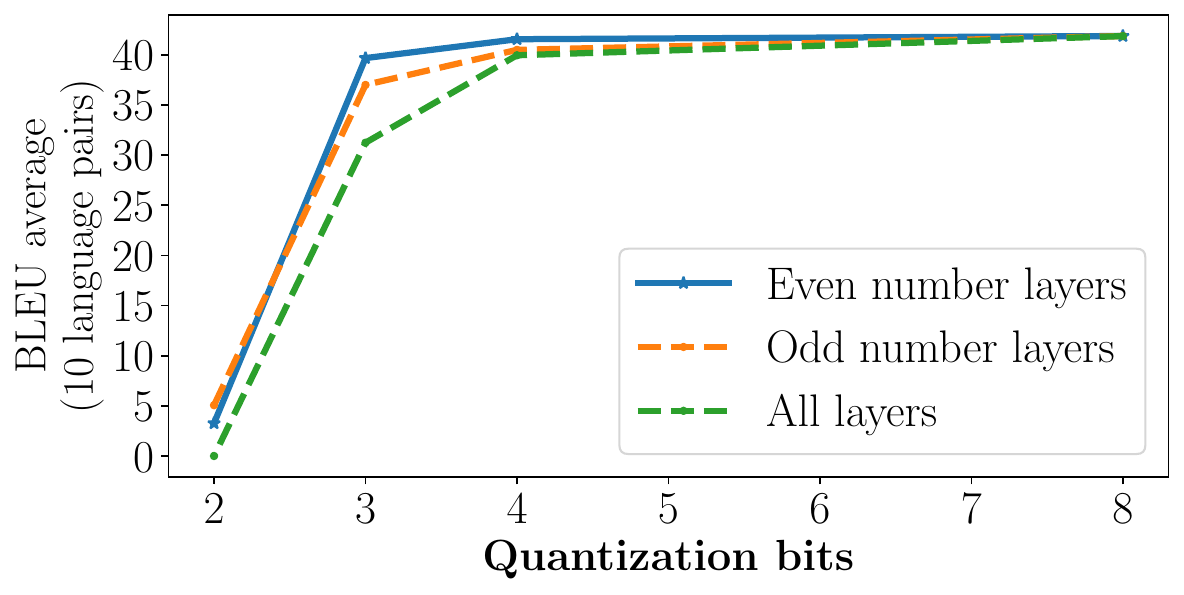}
    \vskip -0.1in
    \caption{Quantization impact of different layers in a dense model.}
    \label{fig:dense-layers}
\end{figure}

For the comparison with MoE models which alternate different block types which are an expert block and a dense block, we consider quantizing only half of the dense transformer blocks' FFNs, because we quantize expert weights only on MoE models which exist only in every other block (even numbered). We compare three different configurations - (1) quantizing even numbered blocks' FFNs only, (2) quantizing odd numbered blocks' FFNs only and (3) quantizing all FFN layers. As can be seen in Figure \ref{fig:dense-layers}, quantizing even numbered blocks' FFNs affects the accuracy the least, and quantizing all FFN layers give the worst result. Based on this experiment, we quantize only even numbered transformer blocks' FFNs for the dense model in all the experiments and comparisons.

\section{Skewness of weight matrices in MoE and dense models}
\label{app:skew}
In the analysis of model weight distribution in Section \ref{sec:quant_methods}, we observe that dense models' FFN layers tend to have more outliers than MoEs' expert FFN layers. We measure the skewness of weight distribution of those in Table \ref{tab:skewness}.

\begin{table}[ht]
\caption{Expert vs non-expert FFN layers parameters distribution skewness}
\label{tab:skewness}
\begin{center}
\begin{small}
\begin{tabular}{cccccccccccc}
\hline
\multicolumn{1}{c}{\bf Parameter} & \multicolumn{1}{c}{\bf skew} \\

\hline
encoder expert 15 FFN fc1 layer 0 & -0.002 \\
\hline
encoder expert 15 FFN fc2 layer 0 &	-0.190 \\ 
\hline
encoder expert 15 FFN fc1 layer 6 & -0.002 \\
\hline
encoder expert 15 FFN fc2 layer 6 &	-0.002 \\
\hline
encoder non-expert FFN fc1 layer 1 & -0.019 \\ 
\hline
encoder non-expert FFN fc2 layer 1	& -10.729 \\ 
\hline
encoder non-expert FFN fc1 layer 7 & 0.003 \\ 
\hline
encoder non-expert FFN fc2 layer 7 & -1.574 \\
\hline
\hline

        encoder expert FFN fc1 mean & 0.00  \\ \hline
        encoder expert FFN fc2 mean & -0.63  \\ \hline
        decoder expert FFN fc1 mean & 0.00  \\ \hline
        decoder expert FFN fc2 mean & 0.48  \\ \hline
        encoder non-expert FFN fc1 mean & 0.00  \\ \hline
        encoder non-expert FFN fc2 mean & -1.84  \\ \hline
        decoder non-expert FFN fc1 mean & 0.00  \\ \hline
        decoder non-expert FFN fc2 mean & -0.09  \\ \hline
\end{tabular}
\end{small}
\end{center}
\end{table}

\section{Machine translation dataset summary}
\label{app:datastat}
Table \ref{tab:datastat} shows the number of parallel sentences used to train dense and MoE models. All languages have at least 300 million sentences and the differences in the number among languages are less than two times.

\begin{table}[ht]
\caption{The number of parallel sentences including backtranslation data.}
\label{tab:datastat}
\begin{center}
\begin{small}
\begin{tabular}{ccccccccc}
\hline
\multirow{2}{*}{\bf Language} & \multicolumn{2}{c}{\bf Number of parallel sentences (million)}        \\
 & \multicolumn{1}{c}{\bf xx $\rightarrow$ English} & \multicolumn{1}{c}{\bf English $\rightarrow$ xx}   \\
\hline
DE (German) & 505 & 411 \\
ES (Spanish) & 448 & 407  \\
FR (French) & 448 & 376  \\
IT (Italian) & 447 & 303  \\
NL (Dutch) & 302 & 378 \\
\hline
\end{tabular}
\end{small}
\end{center}
\end{table}

\section{Robustness comparison between MoE and dense models}
We compared robustness against low-bit quantization between MoE and dense models using the post-training quantization without any QAT. For the dense model, quantization with different bits was applied to the even numbered FFN layers. Appendix \ref{app:dense-layers} shows this is the best layer selection for the dense model. We used two different datasets to verify the proposed quantization method works in different model settings.
\begin{figure}[h]
    \centering
    \includegraphics[width=0.5\textwidth]{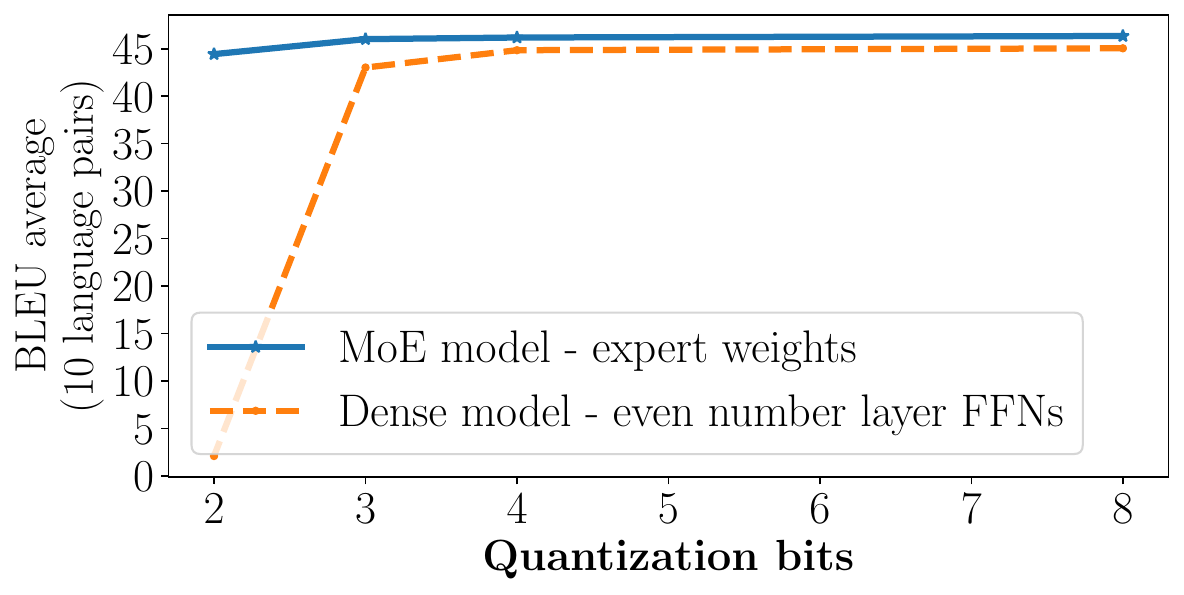}
    \vskip -0.1in
    \caption{Quantization performance comparison between MoE and dense models. 10 different language pair scores are averaged.}
    \label{fig:dense-moe-euro}
\end{figure}

Figure \ref{fig:dense-moe-euro} presents the experiment with the model trained with the larger dataset. It shows the average BLEU scores with different quantization precision for both MoE and dense models. The MoE model can maintain accuracy within -0.3 down to 3-bit and -1.82 for 2-bit. On the other hand, the dense model can preserve the accuracy only down to 4-bit, but starts to lose significant accuracy more than 2 BLEU scores when it goes down to 3-bits. In case of 2-bits, dense model loses most of capability by -42.96 BLEU scores.

\begin{figure}[h]
    \centering
    \includegraphics[width=0.5\textwidth]{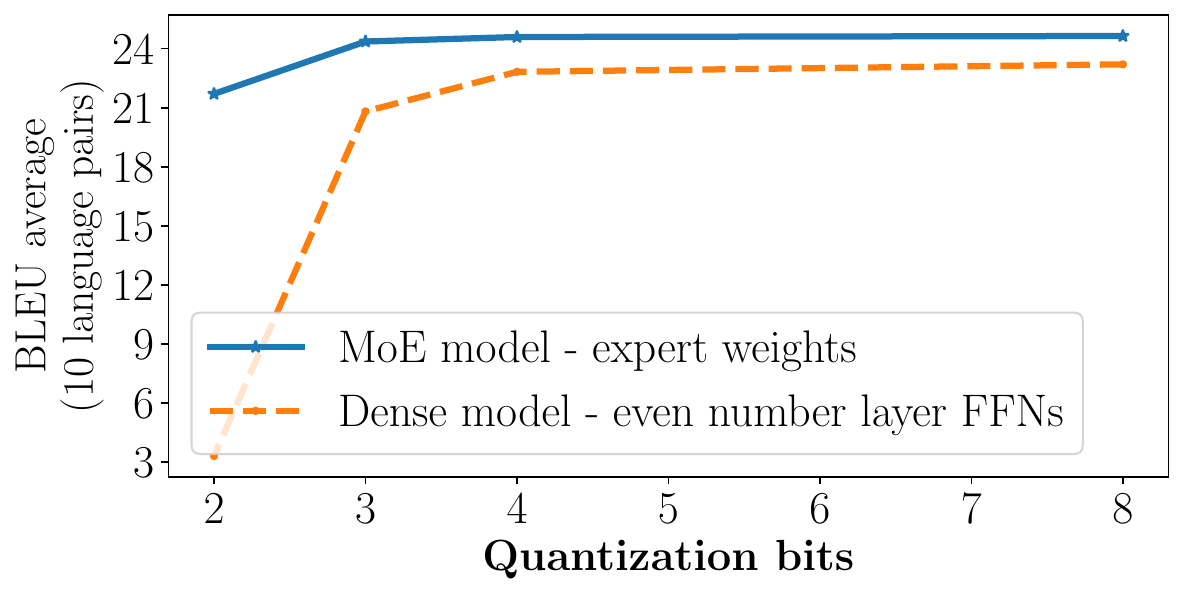}
    \vskip -0.1in
    \caption{Quantization performance comparison between MoE and dense models. 20 different WMT language pairs are averaged.}
    \label{fig:dense-moe-wmt10}
\end{figure}

Figure \ref{fig:dense-moe-wmt10} presents the experiment with the model trained with the smaller dataset. In this setting, each individual expert is smaller, but there are 4 times more experts in one MoE layer. And, they are trained with smaller dataset, so they do not have equivalent knowledge as the previous model trained on the larger dataset. As can be seen in the Figure, the quantization performance shows a similar pattern. The MoE model preserves accuracy even when it is quantized to 2 or 3 bits. However, dense model quickly loses the performance when it is quantized down to lower than 4-bit. Again, the MoE model is much more robust to quantization than the dense model.

\subsection{Robustness of MoE FFN layers to quantization}
\label{sec:sensitivity}
For MoE models, we also conducted a set of experiments with various quantization bits. We divide an MoE model into four parts: (i) expert FFNs, (ii) dense FFN layers, (iii) self-attention layers and (iv) cross-attention layers.

\begin{figure}[h]
    \centering
    \includegraphics[width=0.5\textwidth]{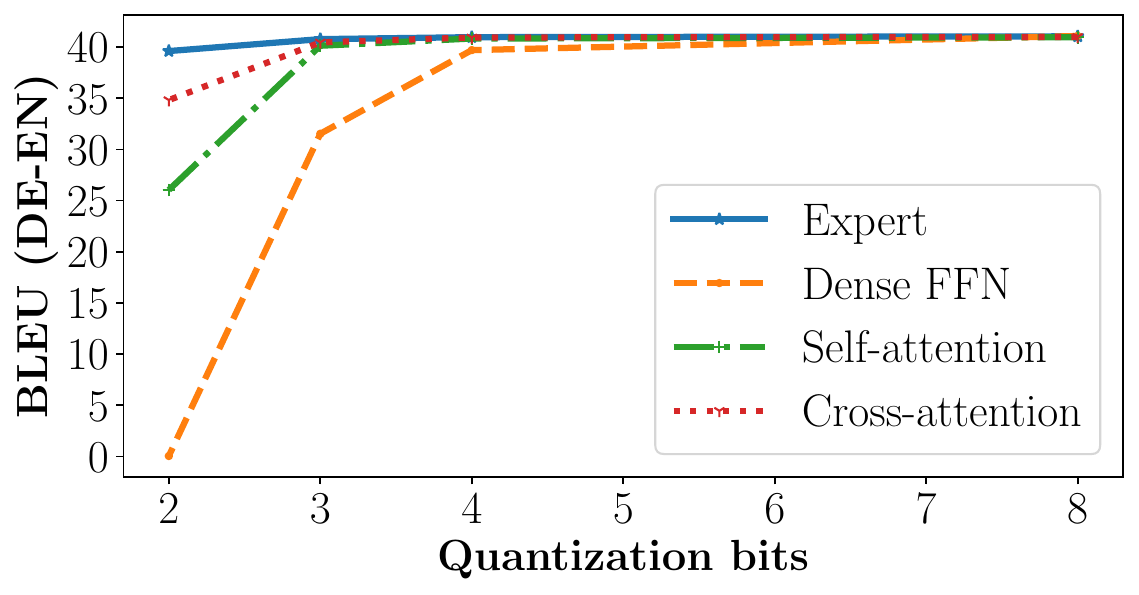}
    \vskip -0.1in
    \caption{Quantization impact on different MoE model parts (channel-wise linear quantiztation without any additional training).}
    \label{fig:different NN parts quant}
\end{figure}

Figure \ref{fig:different NN parts quant} shows the evaluation BLEU scores when different parts of the MoE model are quantized. It is observed that quantizing expert FFN layers to 2-bit does not significantly impact the overall model quality. However, quantizing other parts of the model into 2-bit significantly hurts the output quality. Quantized cross-attention and self-attention blocks can still maintain the quality with 3-bit quantization, but their performance gets impacted with 2-bit quantization. On the other hand, dense FFN layers are significantly impacted by lower bit quantization of 2-bit and 3-bit. With 3-bit quantization, the model score drops by 23 \% of the original score, and 2-bit quantization on dense FFN layers gives almost zero score. The same study is also included on a dense model in Appendix \ref{app:dense-layers}, and a similar pattern with 2 and 3 bit quantization is observed.

\end{document}

%% file: math_commands.tex
\usepackage{amsmath,amsfonts,bm}

\def\eqref#1{equation~\ref{#1}}

\def\1{\bm{1}}

\def\vs{{\bm{s}}}

\def\mA{{\bm{A}}}

\def\mP{{\bm{P}}}
\def\mQ{{\bm{Q}}}

\def\mT{{\bm{T}}}

\DeclareMathAlphabet{\mathsfit}{\encodingdefault}{\sfdefault}{m}{sl}
\SetMathAlphabet{\mathsfit}{bold}{\encodingdefault}{\sfdefault}{bx}{n}

\DeclareMathOperator{\integer}{int}

%% file: figures.tex
\usepackage{pgfplotstable} 

\newcommand\chartOptOneThree {
\begin{tikzpicture}
    \begin{axis}[
        x=1.0cm,
        height = 6cm,
        major x tick style = transparent,
        ybar=2*\pgflinewidth,
        bar width=6pt,
        ymajorgrids = true,
        ylabel style={yshift=-1.5em},
        ylabel = {Speedup over FP16},
        xlabel style={yshift=-0.25em},
        xlabel = {Number of Rows in Activation},
        symbolic x coords={1,8,32,64,128,256,2048,16384},
        xtick = data,
        scaled y ticks = false,
        enlarge x limits=0.07,
        ymin=0,
        legend cell align=center,
        legend style={
            draw=none,
            text depth=0pt,
            at={(0.1,1.0)},
            anchor=south west,
            legend columns=-1,
            column sep=1cm,
            /tikz/column 3/.style={column sep=0pt,font=\bfseries},
            /tikz/every odd column/.append style={column sep=0cm},
        }
    ]
    
        \addplot[style={blue,fill=blue,mark=none}]
            coordinates {(1, 1.88) (8,1.86) (32,1.74) (64, 1.47) (128,1.17) (256,0.98) (2048,0.91) (16384,0.65)};

        \addplot[style={orange,fill=orange,mark=none}]
            coordinates {(1, 2.57) (8,2.54) (32,2.36) (64, 1.55) (128,1.16) (256,0.92) (2048,0.87) (16384,0.64)};

        \addplot[style={black!60!green,fill=black!60!green,mark=none}]
            coordinates {(1, 2.52) (8,2.52) (32,2.32) (64, 1.53) (128,1.19) (256,0.94) (2048,0.86) (16384,0.63)};

        \addplot[red,sharp plot,update limits=false,] coordinates { ([normalized]-1,1) ([normalized]8,1) };

        \legend{INT8, INT4, INT4 (64)}
    \end{axis}
\end{tikzpicture}
}

\newcommand\chartOptThreeZero {
\begin{tikzpicture}
    \begin{axis}[
        x=1.0cm,
        height = 6cm,
        major x tick style = transparent,
        ybar=2*\pgflinewidth,
        bar width=6pt,
        ymajorgrids = true,
        ylabel style={yshift=-1.5em},
        ylabel = {Speedup over FP16},
        xlabel style={yshift=-0.25em},
        xlabel = {Number of Rows in Activation},
        symbolic x coords={1,8,32,64,128,256,2048,16384},
        xtick = data,
        scaled y ticks = false,
        enlarge x limits=0.07,
        ymin=0,
        legend cell align=center,
        legend style={
            draw=none,
            text depth=0pt,
            at={(0.1,1.0)},
            anchor=south west,
            legend columns=-1,
            column sep=1cm,
            /tikz/column 3/.style={column sep=0pt,font=\bfseries},
            /tikz/every odd column/.append style={column sep=0cm},
        }
    ]
        \addplot[style={blue,fill=blue,mark=none}]
            coordinates {(1, 1.71) (8,1.70) (32,1.67) (64, 1.52) (128,0.911) (256,0.94) (2048,0.87) (16384,0.62)};

        \addplot[style={orange,fill=orange,mark=none}]
            coordinates {(1, 2.66) (8,2.64) (32,2.53) (64, 1.46) (128,0.99) (256,0.91) (2048,0.81) (16384,0.65)};

        \addplot[style={black!60!green,fill=black!60!green,mark=none}]
            coordinates {(1, 2.58) (8,2.56) (32,2.42) (64, 1.40) (128,0.95) (256,0.91) (2048,0.80) (16384,0.64)};

        \addplot[red,sharp plot,update limits=false,] coordinates { ([normalized]-1,1) ([normalized]8,1) };

        \legend{INT8, INT4, INT4 (64)}
    \end{axis}
\end{tikzpicture}
}

\newcommand\chartMoeSpeedup {
\begin{tikzpicture}
    \begin{axis}[
        x=1.1cm,
        height = 6cm,
        major x tick style = transparent,
        ybar=2*\pgflinewidth,
        bar width=6pt,
        ymajorgrids = true,
        ylabel style={yshift=-0.8em},
        ylabel = {Speedup over FP16},
        xlabel style={yshift=-0.25em},
        xlabel = {(Batch Size, Beam Width)},
        symbolic x coords={{(1,1)},{(1,2)},{(8,1)},{(8,2)},{(20,1)},{(20,2)},{(32,1)},{(32,2)},{(64,1)},{(64,2)},{(96,1)},{(96,2)}},
        xtick = data,
        scaled y ticks = false,
        enlarge x limits=0.05,
        ymin=1,
        legend cell align=center,
        legend style={
            draw=none,
            text depth=0pt,
            at={(0.35,1.0)},
            anchor=south west,
            legend columns=-1,
            column sep=1cm,
            /tikz/column 3/.style={column sep=0pt,font=\bfseries},
            /tikz/every odd column/.append style={column sep=0cm},
        }
    ]
        \addplot[style={blue,fill=blue,mark=none}]
            coordinates {({(1,1)}, 1.05) ({(1,2)}, 1.04) ({(8,1)}, 1.07) ({(8,2)}, 1.08) ({(20,1)},1.10) ({(20,2)},1.14) ({(32,1)}, 1.12) ({(32,2)},1.15) ({(64,1)}, 1.14) ({(64,2)},1.14) ({(96,1)},1.13) ({(96,2)},1.16)};

        \addplot[style={orange,fill=orange,mark=none}]
            coordinates {({(1,1)}, 1.08) ({(1,2)}, 1.07) ({(8,1)}, 1.12) ({(8,2)}, 1.14) ({(20,1)},1.16) ({(20,2)},1.20) ({(32,1)}, 1.20) ({(32,2)},1.22) ({(64,1)}, 1.23) ({(64,2)},1.24) ({(96,1)},1.22) ({(96,2)},1.24)};

        \legend{INT8, INT4}
    \end{axis}
\end{tikzpicture}
}